# Lazy Propagation in Junction Trees

**Anders L. Madsen**
Department of Computer Science
Aalborg University
Denmark
anders@cs.auc.dk

**Finn V. Jensen**
Department of Computer Science
Aalborg University
Denmark
fvj@cs.auc.dk

## Abstract

The efficiency of algorithms using secondary structures for probabilistic inference in Bayesian networks can be improved by exploiting independence relations induced by evidence and the direction of the links in the original network. In this paper we present an algorithm that on-line exploits independence relations induced by evidence and the direction of the links in the original network to reduce both time and space costs. Instead of multiplying the conditional probability distributions for the various cliques, we determine on-line which potentials to multiply when a message is to be produced. The performance improvement of the algorithm is emphasized through empirical evaluations involving large real world Bayesian networks, and we compare the method with the HUGIN and Shafer-Shenoy inference algorithms.

## 1 Introduction

It has for a long time been a puzzle why "standard" inference algorithms for Bayesian networks did not really use the direction of the links in the network. By "standard" we mean the Lauritzen-Spiegelhalter [Lauritzen and Spiegelhalter, 1988], the Shafer-Shenoy [Shafer and Shenoy, 1990], and the HUGIN [Jensen et al., 1990] algorithms and the various variations over these algorithms ( [Shachter, 1990] and [Jensen, 1995]). These algorithms build a secondary structure (a junction tree or a join tree) by triangulating the (moralized) network. This structure can be used for propagation for all information scenarios. Therefore, the algorithms do not exploit independences induced by the evidence. That is, the tree-structure is large enough to take care of all instantiations of variables. For some (or sometimes all) specific information scenario, a careful exploitation of the $d$-separation properties would result in less complex structures.

Consider for example the Bayesian network indicated in figure 1. If $A$ is instantiated and no evidence has been entered to $DAG_4$, then $DAG_1$, $DAG_2$, and $DAG_3$ are independent, and we need only sent messages down to $DAG_4$. An on-line triangulation of this scenario will result in a much simpler set of junction trees than the off-line produced junction tree. To exploit the specific independences, we need a very efficient algorithm for detecting independences and to perform an efficient triangulation based on these independences. In particular, as the problem of optimal triangulation is $\mathcal{NP}$-complete, there is not much hope that a method requiring on-line triangulation can outperform the "standard" methods for large networks, and improved performance for small networks is not particularly interesting.

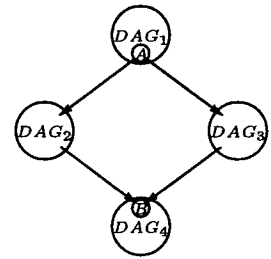

Figure 1: If $A$ is instantiated and no evidence has been entered to $DAG_4$, then it is only necessary to sent messages down to $DAG_4$.

We may relax the requirement to the updating algorithm such that we are only interested in updated probabilities for a very small set of variables. In that case the SPI method [Shachter et al., 1990] and the bucket sort algorithm [Dechter, 1996] can utilize specific independences, as they consist of a collect operation only, where the variables are successively eliminated by multiplying the functions involving $A$ (say) and marginalizing $A$ out of this product. These methods, however, are not able to update all variables efficiently.

In this paper we propose a compromise between off-line triangulation and on-line exploitation of specific independences. We call the method *lazy propagation* as the bulk of the method is lazy evaluation of the potentials for cliques and separators. That is, we work with an off-line produced junction tree, where we have allowed ourselves to use much time on finding a small junction tree. Now, instead of multiplying the conditional probability distributions for the various cliques, we determine on-line which potentials to multiply when a message is to be produced. Thereby, when a message is to be produced, only the required functions are multiplied. An effect of this scheme is that $d$-separation properties induced by evidence are automatically exploited.

The rest of the paper is organized in the following way. Section 2 describes the lazy propagation scheme in detail. In section 3 we present results from a series of real-time tests performed. A discussion of the results is given in section 4, and in section 5 we illustrate how $d$-separation properties are automatically exploited.

## 2 Methods

We briefly review the HUGIN and the Shafer-Shenoy algorithms. For more elaborate presentations, see the references above.

A Bayesian network consists of a graph $G = (\mathcal{V}, \mathcal{E})$ and a probability distribution $P$. $G$ is a directed acyclic graph, $\mathcal{V}$ is the set of variables (which are assumed to be discrete), and $\mathcal{E}$ is the set of edges connecting the variables. The probability distribution $P$ factorizes on $G$ such that:

$$P = \prod_{V \in \mathcal{V}} P(V | pa(V)),$$

where $pa(V)$ is the parent set of $V$. The secondary structures used by the HUGIN and Shafer-Shenoy architectures are constructed from $G$.

A junction tree representation of a Bayesian network $G$ is constructed by moralization and triangulation of $G$. The nodes of the junction tree correspond to cliques of the triangulated graph. A clique is a maximal connected subgraph of the triangulated graph. The cliques of the junction tree are connected by separators such that the so-called junction tree property holds. The junction tree property insures that whenever two cliques $C_i$ and $C_j$ are connected by a path, the intersection, $C_i \cap C_j$ is a subset of every clique and separator on the path. To each clique $C$ and each separator $S$ we associate potentials $\phi_C$ and $\phi_S$, respectively. $\phi_C$ and $\phi_S$ are functions having the variables of $C$ and $S$ as domains. Each variable, $V$, in the Bayesian network has a conditional probability distribution $P(V | pa(V))$. Every distribution is assigned to a clique such that the domain of the distribution is a subset of the clique domain. The set of distributions assigned to a clique, $C$, are combined to form the potential function $\psi_C$. Initially the potentials of the junction tree are given as:

$$\phi_C = \psi_C \quad \text{and} \quad \phi_S = 1$$

for each clique, $C$, and each separator, $S$.

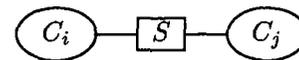

Figure 2: Cliques $C_i$ and $C_j$ are connected by the separator $S = C_i \cap C_j$.

The propagation of evidence in the HUGIN architecture is based on the operation of absorption. Assume $C_i$ and $C_j$ to be neighboring cliques in a junction tree with $S$ as separator, see figure 2. We say that $C_j$ absorbs from $C_i$ if we:

- calculate $\phi_S^* = \sum_{C_i \setminus S} \phi_{C_i}$;

- give $S$ the potential $\phi_S^*$;

- give $C_j$ the potential $\phi_{C_j}^* = \phi_{C_j} \dfrac{\phi_S^*}{\phi_S}$.

The absorption operation is used when a message is sent from one clique to another. Messages flow in two recursive phases, and the flow is controlled by choosing a root clique of the junction tree. The first phase is initiated by collecting evidence to the root and the second phase is initiated by distributing evidence from the root. Collection of evidence to a clique $C$ is done by collecting evidence to all the children of $C$ followed by absorption of evidence from each child. Similarly, distribution of evidence from a clique amounts to absorption of evidence into each child followed by distribution of evidence from the child. After a full round of message passing a message has been sent in each direction along every separator in the junction tree.

The Shafer-Shenoy algorithm can perform inference in a junction tree. The reader should notice that the Shafer-Shenoy algorithm propagates evidence faster in binary join trees than in junction trees [Shenoy, 1997]. The Shafer-Shenoy inference architecture differs from the HUGIN architecture in a number of ways. First, the flow of messages is not controlled by choosing a root of the junction tree. A clique sends asynchronously a message to one of its neighbors when messages from all other neighbors have been received. Second, the clique potentials are not updated during propagation of evidence instead each separator holds two messages. One for each direction. Third, no division of potentials is performed. Consider figure 2 once again. The Shafer-Shenoy message, $\phi_{C_i \to C_j}$, sent from $C_i$ to $C_j$ is calculated as:

$$\phi_{C_i \to C_j} = \sum_{C_i \setminus C_i \cap C_j} \psi_{C_i} \prod_{n \in \mathcal{N}_{C_i} \setminus C_j} \phi_{n \to C_i},$$

where $\psi_{C_i}$ is the clique potential of $C_i$ and $\mathcal{N}_{C_i}$ is the set of neighbors to $C_i$.

After a full round of message passing in the Shafer-Shenoy architecture each separator holds two messages. The clique potential $\phi_{C_i}$ can be obtained by taking the product of all messages sent to $C_i$ and $\psi_{C_i}$.

The message passing scheme for asynchronous firing corresponds to the scheme of CollectEvidence followed by DistributeEvidence. The root is, however, chosen randomly. Let the separator, $S$, between cliques $C_i$ and $C_j$ be the first separator over which messages are sent in both directions, see figure 3. Before $C_j$ sent the message, *new*, over $S$, it has received a message from all its neighbors. This is equivalent to collecting evidence to $C_j$. Sending messages from $C_j$ to all its neighbors is equivalent to distributing evidence from $C_j$.

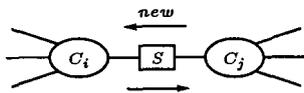

Figure 3: If $C_j$ is the first clique to receive a message from all its neighbors in asynchronous firing, then the messages sent are the same as the messages sent when collecting evidence to $C_j$.

The SPI algorithm and the bucket elimination algorithm do not perform inference based on a secondary structure. Both algorithms are most advantageously used if the reasoning is focused in the sense that only the posterior probability distributions of a small set of target variables are to be calculated. The basic idea behind the SPI and the bucket elimination algorithms is to consider only variables relevant to reasoning about the target set. The variables relevant for a query can be determined from the original Bayesian network by an algorithm which runs in time linear in the number of arcs in the graph. The posterior distribution of the target set is equal to the product of the distributions of the relevant nodes marginalized down to the target set.

## 2.1 Lazy Propagation

The basic idea behind lazy propagation is to take advantage of two important properties of the potentials associated with the nodes of the Bayesian network:

- $\sum_V P(V \mid W) = 1_W$;

- the $d$-separation criterion applies to the potentials.

Instead of combining the probability distributions associated with a clique to obtain the clique potential, we keep the clique potential in factored form, and we change the content of messages passed between cliques in the junction tree. Instead of sending a message consisting of one potential with the set of separator variables as domain, we sent a message consisting of a set of potentials all having domains which are subsets of the separator domain. Messages can flow as in the asynchronous firing scheme or may be controlled by choosing a root as in the HUGIN architecture.

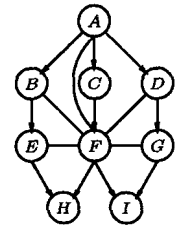

Figure 4: A Bayesian network. Lines without direction are fill ins added during triangulation of the network.

Consider the Bayesian network shown in figure 4 and the corresponding junction tree shown in figure 5. Assume that variable $A$ is instantiated by evidence. Each potential with $A$ in its domain has the domain decreased by $A$. If we assume potentials to be represented as tables, then $P^*$ is the subtable of $P$ corresponding to the instantiation of $A$.

In figure 6 we follow the flow of messages from the leaves of the junction tree towards $ABF$ in the lazy propagation scheme. The message flow corresponds to collecting evidence to $ABF$ in the HUGIN architecture. Assume that the first leaf to sent a message is $EFH$. The potential associated with this clique is $P(H \mid E, F)$. Variable $H$ has to be eliminated, but no calculations are required as $\sum_H P(H \mid E, F) = 1_{EF}$.

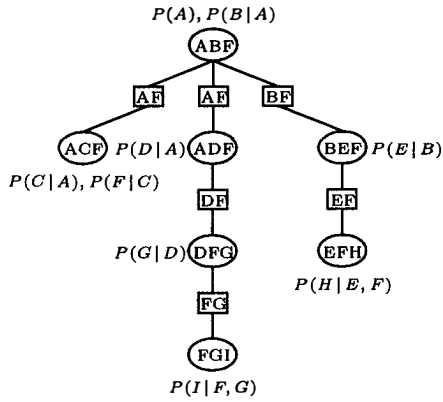

Figure 5: A junction tree constructed from the Bayesian network shown in figure 4. Prior distributions are assigned to cliques as indicated.

The same argument is used when messages are sent from $BEF$, $FGI$, $DFG$, and $ADF$. The last clique to send a message is $ACF$. $ACF$ has $P^*(C)$ and $P(F|C)$ associated, and variable $C$ is eliminated by taking the product of the two potentials and then marginalizing down to $F$.

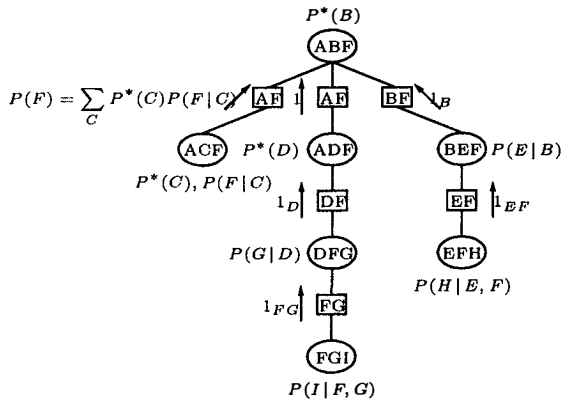

Figure 6: Shows the message flow in the junction tree during the first phase of lazy propagation. The initial messages are sent from the leaves.

At the end of the inward pass, potentials $P^*(B)$ and $P(F)$ are associated with $ABF$. In figure 7 we follow the message flow in the opposite direction. The message sent from $ABF$ to $ACF$ is an empty message as none of the potentials associated with $ABF$ needs to be sent to $ACF$. $P(F)$ is the only potential relevant, but it was sent in the opposite direction during the inward pass of the algorithm (in HUGIN terms it is divided out, in Shafer-Shenoy terms it shall not be transmitted). The message sent from $ABF$ to $ADF$ consists of the potential $P(F)$ as no other potentials are relevant for the subtree rooted at $ADF$. The po-tentials sent from $ADF$ to $DFG$ are $P(F)$ and $P^*(D)$. At $DFG$ we have to combine $P^*(D)$ and $P(G \mid D)$ and marginalize $D$ out to obtain the message to send to $FGI$. $BEF$ is the last child of $ABF$ to receive a message, and the message sent consists of $P^*(B)$ and $P(F)$. Finally, a message containing $P(F)$ and $P(E) = \sum_B P(E|B)P^*(B)$ is sent to $EFH$.

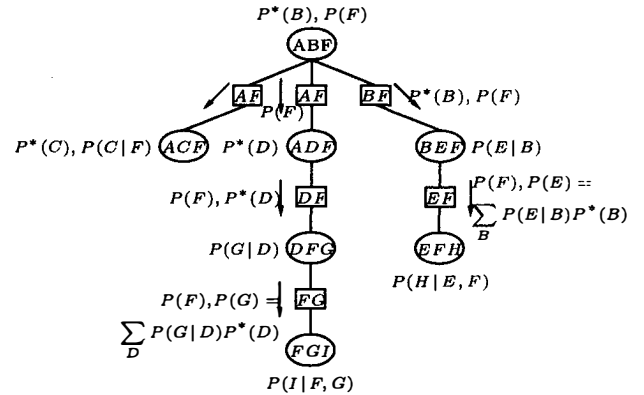

Figure 7: Shows the message flow in the junction tree during the second phase of lazy propagation. The initial messages are sent from $ABF$.

It is not necessary to send the potentials containing only ones. We have included these potentials in the description to make the explanation clear. So, for this example we only performed three marginalizations and all of them involved only two variables.

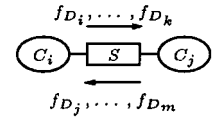

Figure 8: The message sent from $C_i$ to $C_j$ consists of potentials $f_{D_i}, \ldots, f_{D_k}$ and the message sent in the opposite direction consists of $f_{D_j}, \ldots, f_{D_m}$.

The HUGIN architecture imposes a division of separator potentials as described above. The lazy propagation scheme does not require this division, because the combination of potentials is postponed. Consider the two neighboring cliques shown in figure 8, and assume that the message sent from $C_i$ to $C_j$ over the separator $S$ consists of the potentials $f_{D_i}, \ldots, f_{D_k}$ and assume the message sent in the opposite direction to consist of the potentials $f_{D_j}, \ldots, f_{D_m}$. None of the potentials $f_{D_i}, \ldots, f_{D_k}$ are involved in any marginalization when sending from $C_j$ to $C_i$. That is, $f_{D_i}, \ldots, f_{D_k} \subset f_{D_j}, \ldots, f_{D_m}$, and the division operation required in HUGIN propagation quite simply amounts to discarding $f_{D_i}, \ldots, f_{D_k}$ from $f_{D_j}, \ldots, f_{D_m}$. So, lazy propagation dissolves the difference between HUGIN propagation and Shafer-Shenoy propagation.

## 3   Empirical Results

We have tested the lazy propagation scheme to investigate how performance varies with the number of instantiated variables. To get an idea of the performance compared to standard schemes we have implemented Shafer-Shenoy as well as HUGIN propagation. The schemes implemented do only perform propagation. That is, the final step after propagation to marginalize the clique potentials down to each variable is not performed. Also, we have not implemented various speed-up features, like binary join trees or 0-compression.

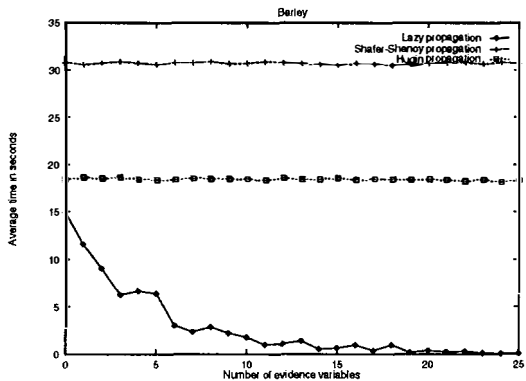

Figure 10: A plot of the average time cost of propagating evidence in the Barley network as a function of the number of variables instantiated.

The tests were performed on a Sun Ultra-2 workstation with two 300 MHz UltraSPARC-1 CPU's running Solaris 2.6 (SunOS 5.6). Each CPU has a 0.5 MB L2 cache. The total RAM on the system is 1024 MB.

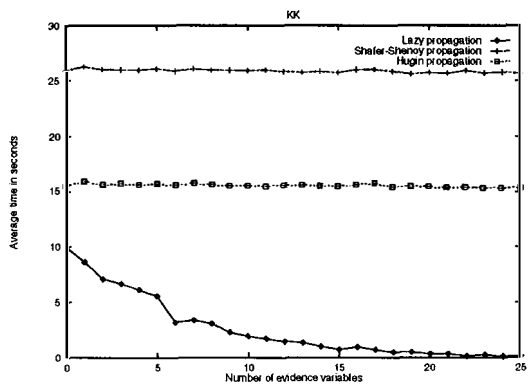

Figure 11: A plot of the average time cost of propagating evidence in the KK network as a function of the number of variables instantiated.

The algorithms were tested on different real-world Bayesian networks with different sizes of evidence sets.

For a given Bayesian network we performed 50 propagations of evidence with the size of the evidence set fixed, but where the evidence variables were chosen at random before each propagation. The number of evidence variables varied from 0 to 50. Figure 9 describes four of the Bayesian networks and corresponding junction trees used for the tests.

Figures 10, 11, 12, and 13 show the average time cost of propagating evidence with the three schemes as a function of the number of variables instantiated in the Barley, the KK, the Diabetes, and the Mildew networks. The figures show that the average time cost of HUGIN propagation for this implementation is always smaller than the average time cost of Shafer-Shenoy propagation, and that the average time cost of lazy propagation decreases considerably as the number of instantiated variables increase.

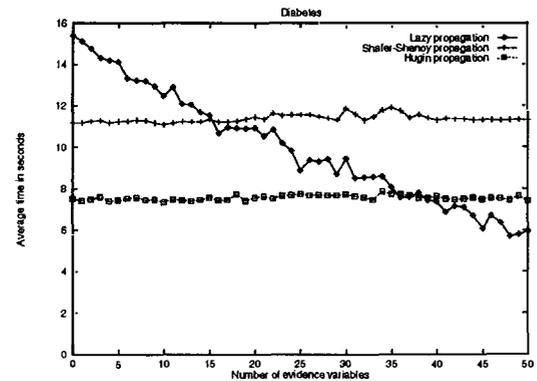

Figure 12: A plot of the average time cost of propagating evidence in the Diabetes network as a function of the number of variables instantiated.

The average time cost of lazy propagation in the Barley, KK, and Mildew networks is smaller than the time cost of the other propagation algorithms even when no variables are instantiated. On average lazy propagation in the Diabetes network becomes faster than Shafer-Shenoy propagation when 16 variables are instantiated, and faster than HUGIN propagation when 39 variables are instantiated.

## 4   Discussion

The experiments indicate that although some evidence may increase time costs, the overall effect of instantiating variables is a decrease of time costs, and with many variables instantiated, lazy propagation outperforms standard propagation schemes.

It seems that time costs are in the same order of magnitude with no instantiated variables for all three

| Network | nodes | node potential size | | | cliques | clique state space size | | | clique neighbors | | |
|---|---|---|---|---|---|---|---|---|---|---|---|
| | | min | max | $\mu$ | | min | max | $\mu$ | min | max | $\mu$ |
| KK | 50 | 2 | 32256 | 2768.1 | 38 | 40 | 5806080 | 397780.2 | 1 | 4 | 1.9 |
| Barley | 48 | 2 | 40320 | 2712.1 | 36 | 216 | 7257600 | 481637.5 | 1 | 4 | 1.9 |
| Diabetes | 413 | 5 | 7056 | 1116.4 | 337 | 495 | 190080 | 30906.3 | 1 | 3 | 2.0 |
| Mildew | 53 | 3 | 280000 | 15633.1 | 29 | 336 | 1756800 | 144133.0 | 1 | 3 | 1.9 |

Figure 9: Information on 4 Bayesian networks and the junction trees generated for these networks.

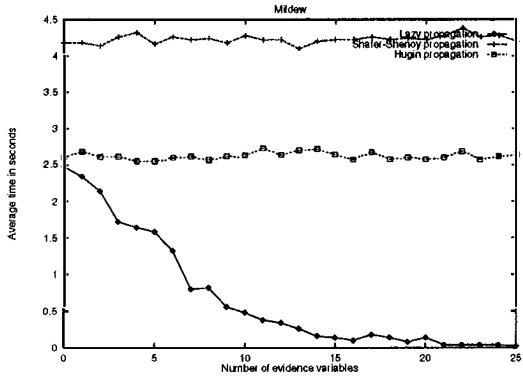

Figure 13: A plot of the average time cost of propagating evidence in the Mildew network as a function of the number of variables instantiated.

schemes, and a rather small set of instantiated variables will yield lazy propagation faster than the standard schemes. Further research is needed to quantify these statements

We have only performed a limited number of tests to investigate how much the space costs are reduced. These tests indicate that the space costs are reduced considerably. This is expected as clique and separator potentials are represented in factored form. Time and space prevents us from giving a thorough elaboration of this topic.

We have done little to speed-up the calculations of a message in the test implementation of lazy propagation. When a message has to be sent from one clique to another some variables have to be marginalized out. If we consider the domain graphs of the potentials relevant to the calculation of the message, then we are faced with a problem similar to the overall problem. That is, we have to calculate the joint probability of a set of nodes in the domain graph. Here any inference algorithm can be used. In the test implementation all relevant potentials are arranged in a list and variables not in the separator domain are eliminated one by one. Variables are eliminated according to the following peeling algorithm:

1. For each variable, $V$, not in the separator domain calculate the domain size of the potential obtained, if $V$ is the next variable eliminated. A variable, $V$, is eliminated by combining all potentials including $V$ in the domain and marginalizing $V$ out.

2. Choose the next variable to eliminate as a variable resulting in the smallest domain size of the potential obtained.

The algorithm is similar to the minimum clique weight heuristic for triangulation, but it is a little different. We do not always eliminate a variable $V$ right away even though its neighbors form a complete graph as in the minimum clique weight heuristic. $V$ is only eliminated right away if there is only one relevant potential containing $V$.

The performance of the lazy propagation scheme depends very much on the topology of the junction tree. If the state spaces of the cliques and separators are large, the lazy evaluation architecture tends to be faster than the other two architectures for small sets of evidence. Sometimes the lazy evaluation architecture is faster even when no variables are instantiated. On the other hand, when the state spaces of the cliques and separators are small, large sets of evidence variables are required before the algorithm becomes faster.

If a Bayesian network has many nodes without parents or many nodes without children, then speed-up is available even when no variables are instantiated. Let $V$ be a variable without parents, then the marginal probability distribution of $V$ can be sent over separators including $V$ right away. Let $W$ be a variable without children and assume that the potential of $W$ is associated with clique $C$. When a message is sent from $C$ over a separator not including $W$ and $W$ has not received evidence, then no calculations are necessary to eliminate $W$ as:

$$\sum_W P(W|pa(W)) = 1_{pa(W)}. \qquad (1)$$

This also applies in the more general case. That is, marginalizing out all head variables of a potential will result in a unity potential with the tail variables as domain.

Some sets of evidence decrease the performance of lazy propagation. Consider the Bayesian network shown in figure 14. A junction tree constructed from this network will contain cliques of the form $D_iE_iF_i$ for $i = 1,\ldots,35$, and these cliques are the only cliques containing $F_i$. No message has to be sent from a $D_iE_iF_i$ clique if variable $F_i$ is not instantiated. If $F_i$ is instantiated, then a message has to be sent from the $D_iE_iF_i$ clique. The lazy evaluation algorithm on average ($n = 50$) uses 4.2 seconds to propagate evidence when no variables are instantiated and 5.4 seconds when variables $F_1,\ldots,F_{35}$ are instantiated.

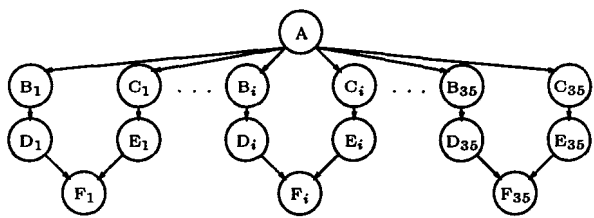

Figure 14: A Bayesian network used to illustrate how evidence might decrease the performance of lazy propagation.

The concept of *barren nodes* was introduced in [Shachter, 1986] and are defined in [Lin and Druzdzel, 1997] as nodes which are neither evidence nor target nodes and have no descendants or only barren descendants. According to this definition no nodes are barren in the lazy evaluation architecture as we are concerned with calculating the posterior probability distribution of all variables in the Bayesian network. The property of barren nodes exploited by algorithms such as the SPI algorithm is that barren nodes have no impact on the posterior probability distribution of the nodes in the target set. This property is exploited in the lazy propagation scheme as described in the next section.

## 5 *d*-separation and Lazy Propagation

Lazy propagation utilizes automatically *d*-separation properties induced by evidence. To illustrate this, consider the Bayesian network, $N$, in figure 15.

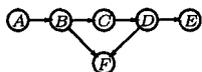

Figure 15: A Bayesian network, $N$, where $A$ and $E$ are independent given $C$.

$N$ has the properties that initially $A$ and $E$ are not *d*-separated. If $C$ is instantiated, then $A$ and $E$ are *d*-separated, but if $C$ and $F$ are instantiated, then $A$ and $E$ are not *d*-separated. In figure 16 a junction tree for $N$ is shown. For the internal elimination order in the cliques we use the peeling algorithm from section 4.

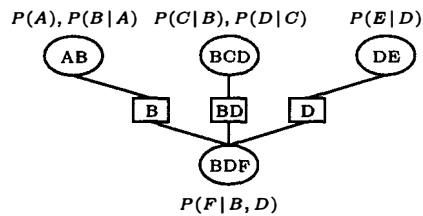

Figure 16: A junction tree for $N$. The potentials associated with the cliques are indicated.

Now, assume that $A$ is instantiated to $a$. In figure 17 we illustrate the flow of potentials towards the clique $DE$. The index of the potentials in the figure indicates the variables relevant for the calculation of the potentials. Index $a$ indicates that the evidence $A = a$ is relevant for the potential.

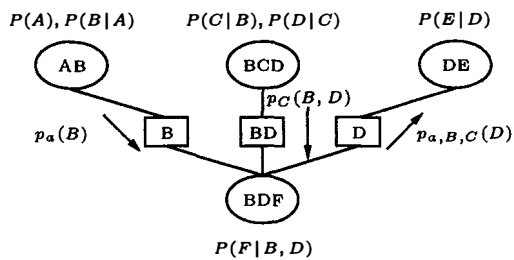

Figure 17: The flow of potentials to the clique $DE$ when $A$ is instantiated to $a$.

As can be seen from figure 17, the evidence $A = a$ is relevant for the updating of $E$. On the other hand, $F$ is irrelevant for $E$, and this has caused a computational saving as the marginalization of $F$ is costless. The cost of propagation is close to the cost of propagating in a junction tree for $N\setminus\{F\}$.

Next, assume that also $C$ is instantiated (to $c$). The flow of potentials is illustrated in figure 18.

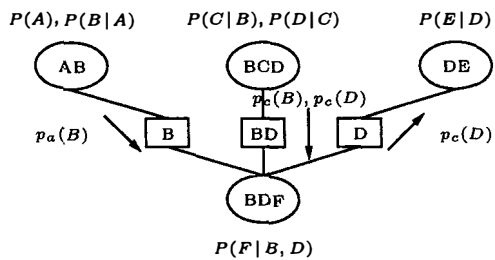

Figure 18: The flow of potentials to the clique $DE$ when $A$ is instantiated to $a$ and $C$ to $c$.

We see that only $C = c$ is relevant for $E$, and the fact that $E$ and $A$ are $d$-separated has yielded substantial savings in the computation. No marginalizations are performed in the propagation.

For completion we illustrate what happens when we furthermore instantiate $F$ to $f$. Then $A$ and $E$ are not $d$-separated, and the resource requirements for updating $DE$ increase.

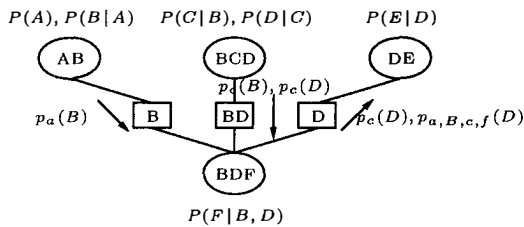

Figure 19: The flow of potentials to the clique $DE$ when $A$ is instantiated to $a$, $C$ to $c$, and $F$ to $f$.

## 6 Conclusion

In this paper we presented an algorithm for probabilistic inference in Bayesian networks. The algorithm exploits the independences induced by evidence and the direction of the links in the original graph. The performance depends on the topology of the original Bayesian network and the junction tree constructed from it.

The test results show that the algorithm performs inference faster than both the HUGIN and the Shafer-Shenoy algorithms if the size of the set of evidence variables is large enough. It should, however, be emphasized that the performance of the test implementations of the Shafer-Shenoy and the HUGIN architectures can be improved by exploiting existing techniques for speeding up the algorithms. Most of these techniques also apply to the lazy evaluation architecture.

The lazy propagation scheme enlarges the class of tractable Bayesian networks as the space costs of this scheme are smaller than the space costs of the HUGIN and Shafer-Shenoy architectures.

**Acknowledgment**

Thanks to the anonymous referees for productive remarks and to the DINA-group at Aalborg University (http://www.cs.auc.dk/research/DSS/DINA).